%% file: main.tex
\documentclass{article}

\usepackage{microtype}
\usepackage{graphicx}
\usepackage{subfigure}
\usepackage{booktabs} 

\usepackage{hyperref}

\usepackage[accepted]{icml2025}

\usepackage{amsmath}
\usepackage{amssymb}
\usepackage{mathtools}
\usepackage{amsthm}

\usepackage[capitalize,noabbrev]{cleveref}

\usepackage{multirow}
\usepackage{xspace}
\newcommand{\name}{\textsc{TransientTrack}\xspace}

\theoremstyle{plain}

\theoremstyle{definition}

\theoremstyle{remark}

\icmltitlerunning{TransientTrack: Advanced Multi-Object Tracking}

\begin{document}

\twocolumn[
\icmltitle{TransientTrack: Advanced Multi-Object Tracking and Classification of Cancer Cells with Transient Fluorescent Signals}



\icmlsetsymbol{equal}{*}

\begin{icmlauthorlist}
\icmlauthor{Florian Bürger}{bioinfo,matnat,cmmc}
\icmlauthor{Martim Dias Gomes}{bioinfo,cmmc}
\icmlauthor{Nica Gutu}{gen,charite}
\icmlauthor{Adrián E. Granada}{charite}
\icmlauthor{Noémie Moreau}{bioinfo,cmmc}
\icmlauthor{Katarzyna Bozek}{bioinfo,cmmc,cecad}
\end{icmlauthorlist}
\icmlaffiliation{bioinfo}{Institute for Biomedical Informatics, Faculty of Medicine and University Hospital Cologne, University of Cologne, Cologne, Germany.}
\icmlaffiliation{matnat}{Faculty of Mathematics and Natural Sciences, University of Cologne, Cologne, Germany.}
\icmlaffiliation{cmmc}{Center for Molecular Medicine Cologne (CMMC), Faculty of Medicine and University Hospital Cologne, University of Cologne, Cologne, Germany.}
\icmlaffiliation{cecad}{Cologne Excellence Cluster on Cellular Stress Responses in Aging-Associated Diseases (CECAD), University of Cologne, Cologne, Germany}
\icmlaffiliation{charite}{Charité Comprehensive Cancer Center, Charité Universitätsmedizin, Berlin, Germany.}
\icmlaffiliation{gen}{Centre for Genomic Regulation, Barcelona, Spain.}

\icmlcorrespondingauthor{Florian Bürger}{florian.buerger@uni-koeln.de}

\icmlkeywords{Machine Learning, ICML}

\vskip 0.3in
]



\printAffiliationsAndNotice{}  

\begin{abstract}
\input{chapter/abstract}
\end{abstract}

\input{body}

\section{Acknowledgments}
Florian Bürger and Katarzyna Bozek were supported by the North Rhine-Westphalia return program (311-8.03.03.02-147635), Martim Dias Gomes was supported by the BMBF program Junior Group Consortia in Systems Medicine (01ZX1917B). Noémie Moreau was supported by the Deutsche Forschungsgemeinschaft (DFG, German Research Foundation) - Project No. 386793560. Florian Bürger, Katarzyna Bozek, Martim Dias Gomes, and Noémie Moreau were hosted by the Center for Molecular Medicine Cologne. We thank the IT Center of the University of Cologne (ITCC) for providing support and computing time.

\bibliography{literature.bib}
\bibliographystyle{icml2025}

\end{document}

%% file: chapter/abstract.tex
Tracking cells in time-lapse videos is an essential technique for monitoring cell population dynamics at a single-cell level. Current methods for cell tracking are developed on videos with mostly single, constant signals and do not detect pivotal events such as cell death. Here, we present TransientTrack, a deep learning-based framework for cell tracking in multi-channel microscopy video data with transient fluorescent signals that fluctuate over time following processes such as the circadian rhythm of cells. By identifying key cellular events – mitosis (cell division) and apoptosis (cell death) our method allows us to build complete trajectories, including cell lineage information. TransientTrack is lightweight and performs matching on cell detection embeddings directly, without the need for quantification of tracking-specific cell features. Furthermore, our approach integrates Transformer Networks, multi-stage matching using all detection boxes, and the interpolation of missing tracklets with the Kalman Filter. This unified framework achieves strong performance across diverse conditions, effectively tracking cells and capturing cell division and death. We demonstrate the use of TransientTrack in an analysis of the efficacy of a chemotherapeutic drug at a single-cell level. The proposed framework could further advance quantitative studies of cancer cell dynamics, enabling detailed characterization of treatment response and resistance mechanisms. The code is available at \underline{https://github.com/bozeklab/TransientTrack}.

%% file: body.tex
\input{chapter/introduction}
\input{chapter/related_work}
\input{chapter/method}
\input{chapter/evaluation}

\input{chapter/analysis}
\input{chapter/conclusion}

%% file: chapter/introduction.tex
\section{Introduction}
\label{sec:introduction}

Multi-Object Tracking (MOT) methods enable observing cell cultures at a single-cell level. Accurate tracking of such systems is essential for analyzing cell lineages, motility, and spatial interactions under various conditions. Compared to classical MOT tasks, cell tracking poses unique challenges. Cells may divide or die, undergo substantial morphological changes (e.g., variations in size, shape, and texture), frequently appear in dense, suboptimal staining with minimal inter-cellular shape variation \cite{akram2017cell}. In particular, low signal-to-noise ratios and low contrast make object detection challenging, as objects become nearly indistinguishable from the background, which further complicates tracking. 

In preclinical cancer research, understanding how individual cells respond to drug perturbations depends on identifying diverse fate decisions, including proliferation, quiescence, and apoptosis. Time-lapse fluorescence microscopy enables direct observation of such behaviors, creating opportunities to quantify therapeutic effects at scale. Detecting cell events like cell death is central for understanding treatment responses. Nevertheless, identifying these events typically relies on manual inspection or highly customized workflows.

For over a decade, state-of-the-art methods have been benchmarked within the Cell Tracking Challenge (CTC) \cite{mavska2023cell}. However, existing CTC datasets primarily consist of single-channel images with constant intensities, emphasizing cell division while neglecting the equally critical event of cell death \cite{scherr2020cell,gallusser2024trackastra,ben2022graph}. 
This imbalance limits the ability to evaluate methods in applications such as quantifying responses to chemotherapeutic drugs, where accurate identification of death events is critical. In addition, many biological studies use multiple imaging channels to probe molecular functions, dynamics, and phenotypes, while these channels often exhibit oscillatory or transient signals that further increase tracking complexity.

To address the lack of apoptosis detection and the challenge of more difficult multi-channel recordings in MOT, we introduce \name, a deep learning–based framework that efficiently tracks cells in multi-channel microscopy data containing transient signals, while also detecting key events of mitosis and apoptosis. Unlike existing methods, which often rely on extensive feature engineering \cite{moen2019accurate,ben2022graph}, our framework performs matching based solely on spatial information and cell detection network-derived embeddings. These embeddings inherently encode essential object visual and spatial features, making them highly suitable for robust and efficient detection association. By directly utilizing the detection embeddings without additional processing, our method remains lightweight, computationally efficient, and well-suited for large-scale time-lapse microscopy experiments.

To evaluate our approach we used a particularly challenging dataset of $309$ time-lapse videos of human osteosarcoma-derived cancer cell-line (U-2 OS) in culture. Each video is $93$ hours long and the cells are tagged with three fluctuating signals, corresponding to the circadian rhythm, cell fitness, and cell cycle. In $117$ videos, cisplatin, a chemotherapeutic drug, was added to induce DNA damage. For a subset of these videos, we manually annotated cell positions, key cellular events (mitosis and apoptosis), and lineage relationships by linking each cell to its parent and daughter cells. After establishing the tracking performance of our method on a subset of annotated videos, we applied it to the full dataset to conduct a large-scale analysis revealing dose-dependent reductions in proliferation, delayed apoptosis onset, and characteristic lineage signatures reflecting how treatment disrupts cell-state inheritance. Together, these results demonstrate how our approach can uncover both global and lineage-specific dynamics in complex imaging datasets.

\textbf{Our main contributions can be summarized as follows}:
\begin{itemize}
    \item We propose a lightweight, two-stage deep learning framework for robust tracking of cells in multi-channel microscopy videos with transient signals. The model jointly detects cells and classifies them as living or dead, integrating biologically meaningful endpoints. 
    \item Our tracking strategy relies solely on learned feature embeddings and spatial cues, without requiring hand-crafted heuristics and continuity ensured through Kalman Filter–based interpolation.
    \item We release a largely annotated dataset of 309 multi-channel fluorescence microscopy videos capturing cell cultures at single-cell resolution.
    \item We demonstrate the utility of our approach in a large-scale study, tracking over $28{,}000$ cell trajectories and quantifying proliferation, apoptosis, and lineage behavior under varying cisplatin dosages. Our method enables single-cell level study of factors that determine cell fate across cell generations and lineages.
\end{itemize}

%% file: chapter/related_work.tex
\section{Related Work}
\label{sec:relatedWork}
Cell tracking is as a subfield of Multi-Object Tracking (MOT), but poses domain-specific challenges that distinguish it from classical MOT tasks. While MOT methods typically focus on handling occlusion, motion blur, and appearance changes of objects \cite{wang2024smiletrack}, cell tracking must additionally account for biological events such as mitosis and apoptosis. A common approach for MOT is the tracking-by-detection paradigm, which separates the process into two distinct steps: object detection and object identity association. This modular design allows for independent optimization and enhancement of each component. While the detection step is typically performed with deep learning models, the association step often relies on heuristics and algorithmic solutions \cite{bergmann2019tracking,stanojevic2024boosttrack,advzemovic2025deep,dai2022survey}.

\subsection{Multi-Object Tracking}
\label{subsec:relatedWork:MOT}
In many MOT approaches, association techniques, like IoU Tracker \cite{bochinski2017high} matches bounding boxes across frames based on IoU values using the Hungarian Algorithm \cite{kuhn1955hungarian}. Others associate bounding boxes based on IoU values with greedy matching \cite{yang2021multi}. However, IoU can fail in scenarios like cell division, where changes in shape or size reduce overlap. In such cases, centroid-based distances are more robust. Instead of the IoU values, SORT \cite{bewley2016simple} predicts tracklet positions with a Kalman Filter \cite{kalman1960new} and uses the spatial distances between objects and predictions. DeepSORT \cite{wojke2017simple} combines object distances and appearance information into the matching to also account for re-identification of objects in complex tracking scenarios. It combines the Mahalanobis distance \cite{fukunaga2013introduction} alongside the cosine distance of the appearance feature embeddings of the detected objects. Alternatively, CenterPoint \cite{yin2021center} employs $L^2$ distance between centroids to improve matching reliability. To address challenges posed by hard-to-detect objects, ByteTrack \cite{zhang2022bytetrack} improves tracking robustness by incorporating both high-confidence and low-confidence detections into a two-stage association strategy. The first stage involves matching high-confidence detections to existing tracks and the second associating unmatched tracks with low-confidence detections. This approach has inspired several advancements, including BOT-SORT \cite{aharon2022bot}, Fast OSNet \cite{li2024lightweight}, RLM-Tracking \cite{ren2024rlm}, and DeconfuseTrack \cite{huang2024deconfusetrack}, achieving state-of-the-art performance across diverse tracking challenges.

\subsection{Cell Tracking}
\label{subsec:relatedWork:cellTracking}
Recent advancements in cell tracking utilize graph-theory approaches, which primarily focus on modeling associations between cell observations rather than performing detection themselves \cite{chatterjee2013cell,scherr2020cell,ben2022graph}. In these frameworks, candidate graphs are built to represent possible cell trajectories, which are then resolved using optimization strategies such as minimum-cost flow \cite{padfield2011coupled}, Integer Linear Programming (ILP) \cite{turetken2016network,gallusser2024trackastra,akram2017cell,padfield2011coupled}, or dynamic programming methods like the Viterbi algorithm \cite{magnusson2014global}. Variants of these methods include using phase correlation between cells as an association cost or introducing edges for lost cells to enable re-identification \cite{scherr2020cell}. Other approaches, such as \cite{gallusser2024trackastra}, explicitly assume annotated detections as given and employ a transformer to learn association costs from object features before resolving the graph via greedy linking or discrete optimization. By focusing on association rather than detection, graph-based methods aim to reduce the influence of detection errors (false positives and false negatives) on downstream tracking. Nevertheless, they often leave challenges such as bridging gaps due to missing detections, detecting apoptosis, or integrating multiple transient signals unaddressed. In general, \cite{akram2017cell} argues that such graph-based approaches introduce additional complexities (e.g., ILP is NP-hard) and may still fail to rectify detection errors.

In contrast, approaches like the MU-US team in the Cell Tracking Challenge \cite{mavska2023cell} ensure consistent tracks by recovering occluded or missed detections via interpolating positions with a Kalman Filter \cite{kalman1960new}. However, they do not perform classification nor detect specific events like apoptosis. Furthermore, existing methods, especially based on CTC data, typically assume stable illumination and pixel intensities, and --- to the best of our knowledge --- none explicitly handles the challenge of integrating multiple transient, non-constant signals alongside apoptosis detection.

%% file: chapter/method.tex
\section{Methodology}
\label{sec:method}
\subsection{Data}
\label{subsec:method:data}
Our dataset examines the proliferation status and cell cycle phase response of single cells to a chemotherapeutic drug cisplatin \cite{granada2020effects}. The dataset comprises $309$ videos, each containing $234$ frames at a size of $1024 \times 1024$ pixels per frame. Frames were captured at $30$-minute intervals over a total duration of $117$ hours per video using a fluorescence time-lapse microscopy setup. To monitor cellular dynamics, three distinct fluorescent markers --- cell cycle reporter hGeminin-CFP, circadian clock reporter NR1D1::VNP, and cell fitness reporter p53-mKate --- were employed, resulting in multi-channel images that capture transient, oscillatory signals over time, providing insight into the state and activity of cells. In $118$ of these videos, cisplatin was added after $48$ hours (frame $96$) in three dosage groups: high ($46$ videos, $13$ µM), medium ($36$ videos, $10$ µM), and low ($36$ videos, $7$ µM). Cisplatin affects DNA synthesis and DNA-damage repair, leading to cell division inhibition and ultimately cell death \cite{yu2016cisplatin}. We also have $36$ control videos where cisplatin was not added. 

We annotated cell trajectories, mitotic events, and apoptotic events across a subset of the videos. We labeled $274$ videos for $186$ frames. Following cell division, daughter cell identifiers preserve lineage information. The labeled dataset contains $51{,}000$ cell tracks, averaging $187$ tracks per video. Across the dataset, $18{,}000$ cell divisions and $3{,}500$ cell deaths were marked in total. From this subset, we only excluded one video from the the medium-dose group.

We utilized $10$ of the annotated videos to evaluate tracking performance. The remaining annotated videos were used to train the detector. They were randomly split into training, validation, and test sets with a $0.8$/$0.1$/$0.1$ ratio, respectively. For analysis, we utilized a subset of the whole dataset for which we have drug dosage information ($154$ videos), including all $234$ frames per video. We published the complete dataset at \underline{https://doi.org/10.4126/FRL01-006526608}.

\subsection{Cell Tracking Pipeline}
\label{subsec:method:pipeline}
\begin{figure*}[t]
    \centering
    \includegraphics[width=0.8\textwidth]{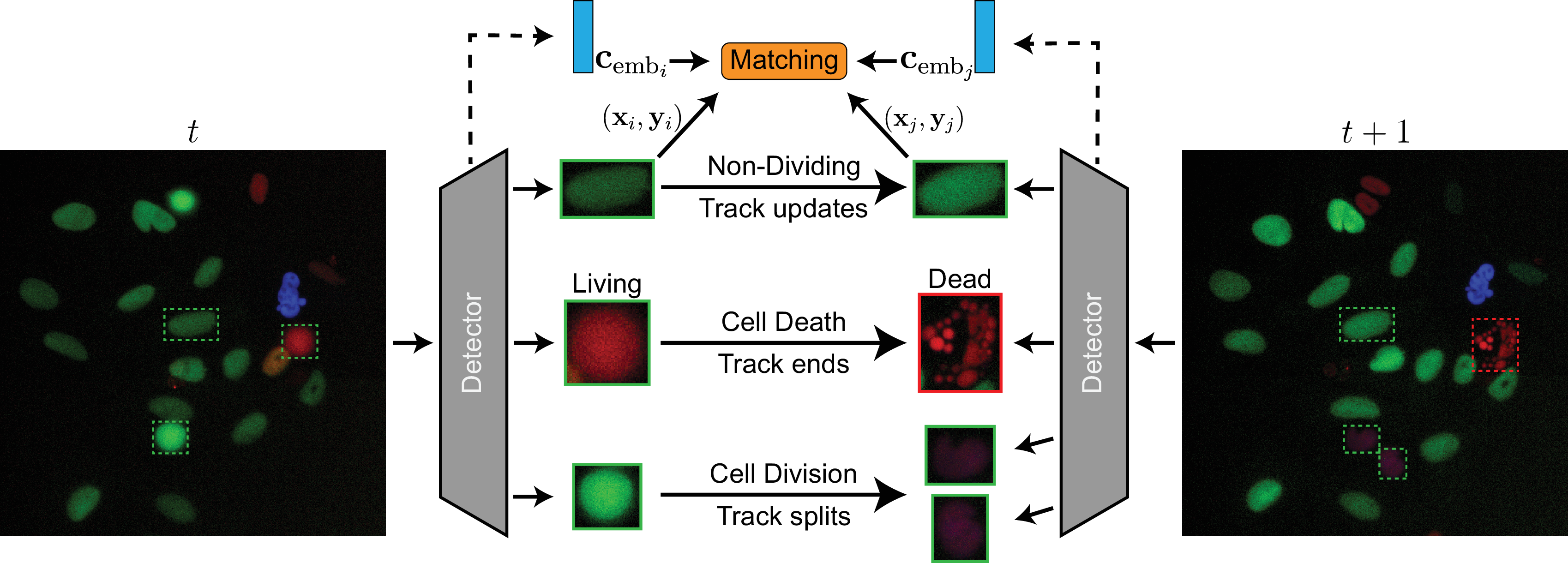}
    \caption{\name\ is a deep learning-based tracking method that utilizes Deformable DETR as a detector. It tracks cells across consecutive frames and identifies cellular events such as mitosis and apoptosis by leveraging object embeddings from the decoder along with spatial information.}
    \label{fig:pipeline_overview}
\end{figure*}

\name is a deep learning-based cell tracking method in live-cell microscopy video data featuring multiple distinct transient signals. An overview of the pipeline is illustrated in Fig.~\ref{fig:pipeline_overview}. Our approach enables robust cell tracking across multiple channels while recognizing key cellular events - mitosis and cell death. \name follows the tracking-by-detection paradigm. In the first step, all cells are detected independently across consecutive frames by processing the raw multi-channel images with transient signals. In the second step, detections from subsequent frames are matched to construct continuous trajectories. 

Our approach combines several recent advancements in deep learning, MOT, and cell tracking (e.g., Transformer Networks, the utilization of all detection boxes in a two-stage matching algorithm), with proven tracking techniques such as Kalman Filter-based interpolation. This integrated design yields accurate and continuous trajectories, even under challenging image conditions.
Our approach integrates several recent advancements in deep learning, MOT, and cell tracking, including Transformer Networks, the utilization of all detection boxes in a two-stage matching algorithm, and the interpolation of missing tracklets with the Kalman Filter. This unified framework ensures accurate and continuous cell trajectories, even under challenging imaging conditions. 

\subsubsection{Detection \& Classification}
\label{subsubsec:method:pipeline:detection}
For detection, we employ Deformable DETR \cite{zhu2020deformable}. The model consists of a CNN as backbone for feature extraction, followed by a Transformer encoder–decoder architecture. The decoder output is passed to two prediction heads: a classification head, which distinguishes between alive and dead cells, and a regression head, which estimates bounding boxes for the detected cells. The matching process relies solely on spatial cell information and Transformer decoder embeddings, which inherently capture the object representations used by prediction heads for bounding box localization and classification. By directly using these decoder outputs, our method remains lightweight and efficient, avoiding any additional tracking-specific feature computation. Deformable DETR outputs candidate bounding boxes with associated confidence scores and class labels (living or dead cell). Given the challenges of detecting cells with low-intensity signals and background noise, we leverage the ByteTrack \cite{zhang2022bytetrack} approach to include low-confidence detections in a two-stage association process. Similarly to \cite{aharon2022bot,li2022simpletrack,wang2024smiletrack}, detections are first split into high-confidence and low-confidence sets using a threshold $\tau$, where detections below a lower bound $\tau_{\text{low}}$ are discarded. High-confidence detections are treated as reliable cell candidates and are prioritized for track initialization, with those from the first frame establishing the initial set of tracks. Each detection is represented by a bounding box and a classification label (living or dead).

\subsubsection{Matching}
\label{subsubsec:method:pipeline:matching}
Our matching is carried out sequentially between two consecutive frames. During the first stage of matching, we match high-confidence detections of frame $t$ and $t+1$ with each other. We base our matching on the distance between the centroids of the detections and the similarity of their embeddings. For the distance, we use the Euclidean distance $d(c,\hat{c})$ which is the distance between the centroids $c_{\text{cent}}$ and $\hat{c}_{\text{cent}}$ of two cells $c$ and $\hat{c}$, respectively. To enforce embedding consistency during matching, we introduce a similarity threshold $\tau^{\text{sim}}$ applied to the $L^1$ distance between cell embeddings $c_{\text{emb}}$ and $\hat{c}_{\text{emb}}$:
\begin{equation}
\label{eq:sim}
    \text{sim}(c,\hat{c}) = || c_{\text{emb}} - \hat{c}_{\text{emb}} ||_{1}
\end{equation}
In cell tracking, we deal with a one-to-many matching scenario: a cell in frame $t$ can either be associated with a single cell in frame $t+1$, or with multiple cells in frame $t+1$, in which case we assume that the cell has divided. To ensure this, we must eliminate detections that appear in multiple sets of possible candidates. To resolve such conflicts, we retain the candidate that minimizes the following weighted sum: 
\begin{equation}
\label{eq:del_candidates}
   \lambda \cdot \frac{\text{sim}(c, \hat{c})}{\tau^{\text{sim}}} + (1-\lambda) \cdot \frac{d(c, \hat{c})}{\tau^{\text{dst}}},
\end{equation}
Similar to DeepSORT \cite{wojke2017simple}, we introduce a weighting parameter $\lambda$ with $\lambda\in[0,1]$. 

After resolving conflicts, a cell $c$ in frame $t$ can have three possible outcomes:
\begin{enumerate}
   \item \textbf{No Candidates}: If no viable match is found, $c$ is included in the second round of association.
   \item \textbf{Single Candidate}: If exactly one match is identified, the corresponding track is updated with the candidate's information.
   \item \textbf{Multiple Candidates}: If multiple candidates remain, we assume a mitotic event. In this case, $c$ is marked as complete, new tracks are initialized for the daughter cells, and $c$ is indicated as their mother cell.
\end{enumerate}

To detect cell death, we match a living cell $c$ in frame $t$ to a cell $\overline{c}$ in frame $t+1$ classified as dead, again using the weighted sum equation (\ref{eq:del_candidates}). Since dead cells typically differ in appearance from living ones, similarity scores are generally low, but spatial proximity remains high. Such a match marks the end of the cell’s lifespan and lineage. From that point onward, $\overline{c}$ can only be associated with other detections classified as dead, ensuring that once a cell is marked as dead, it cannot reappear as living in subsequent frames.

Unmatched high-confidence detections from frame $t$ are retained for up to $n$ frames in a memory bank and considered in matchings in following frames. They are kept for only a limited number of matching attempts to avoid false matchings. Due to their movement the cells are unlikely to remain in the same location in following frames. 

In the second matching stage, we use low-confidence detections and match them with the remaining unmatched trajectories. As the detector may assign low confidence to cells due to low transient signals or background noise, we incorporate these detections into a re-identification process and match them to the stored, previously tracked cells. For re-identification, we also employ the Kalman Filter \cite{kalman1960new} to interpolate missing frames within the tracks. We initialize the Kalman Filter with the first known position of the lost cell and then update it with every subsequent position, allowing it to learn the motion dynamics of the cell. This way, we can interpolate the track's missing positions with the Kalman Filter's predictions. This approach ensures continuous trajectories by compensating for frames in which the cells were missed. Furthermore, continuous tracks are crucial for evaluation using the CTC metrics. 

\subsection{Training}
\label{subsec:method:training}
We only trained the Deformable DETR component, as our association technique is not deep learning–based. Both the CNN backbone and the encoder–decoder Transformer were trained from scratch, following the procedure described in \cite{zhu2020deformable} and incorporating iterative bounding box refinement. Training Deformable DETR involves a combination of three losses. For classification, we used a binary focal cross-entropy loss \cite{lin2017focal}, which down-weighs easy negatives and focuses the optimization on hard or misclassified examples. Bounding box regression was guided by two complementary losses: the $\ell_1$ loss, which penalizes deviations in box coordinates, and a Generalized IoU (GIoU) loss \cite{rezatofighi2019generalized}, which accounts for both overlap and spatial alignment of predicted and ground-truth boxes, improving robustness when boxes do not overlap. Each loss is weighed by a coefficient. We adopted the coefficients from the original Deformable DETR training, except for the classification loss, which we increased to $3$ to place greater emphasis on accurate class prediction. The model was optimized with AdamW using separate learning rates $0.0001$ for the CNN backbone and $0.001$ for the Transformer encoder–decoder, a weight decay of $0.0001$, and a batch size of $4$. Early stopping was applied, terminating training after $50$ epochs without improvement. Data augmentation included horizontal and vertical flipping, Gaussian blurring, and color jittering. All experiments were performed on a system equipped with an Intel Xeon Gold 5218 CPU, 150 GB of RAM, and an NVIDIA Tesla V100-SXM2-32GB GPU.

%% file: chapter/evaluation.tex
\section{Experiments \& Results}
\label{sec:evaluation}
\subsection{Evaluation Setup}
\label{subsec:evaluation:setup}

To evaluate our approach, we used the official CTC metrics \cite{mavska2023cell,matula2015cell}, which are widely used for benchmarking cell tracking methods. Specifically, we assessed performance using 1) the Detection Accuracy (DET) metric, which quantifies how well individual cells have been identified in each frame, 2) the Linkage Accuracy (LNK) metric, which focuses solely on the correctness of temporal associations between cells, independently of their detection accuracy, and 3) the Tracking Accuracy (TRA) metric, which evaluates both, the correct identification of cells and the consistency of their successive labeling throughout the sequence. All three metrics lie in the range $[0,1]$, where higher values indicate better performance, and they are based on the acyclic oriented graph matching (AOGM) measure \cite{matula2015cell}, which quantifies the number of transformations required to transform a predicted detection or tracking graph into the respective ground truth graph. AOGM penalizes errors such as false positives, false negatives, missing links, and incorrect links. The LNK score isolates only the penalties related to link errors, assuming detections are correct, thus providing a direct assessment of local temporal consistency in tracking. In contrast, for the calculation of TRA, missing links, wrong links, and links with wrong semantics are considered in addition to the detection errors. Consequently, TRA measures the global temporal consistency. In brief, the DET score includes only detection errors, LNK evaluates exclusively the correctness of cell-to-cell links across frames, and TRA incorporates both detection and tracking errors.

Alongside the CTC metrics, we employed MOT measures. Although these metrics do not capture biological events, they provide a comprehensive assessment of detection and tracking performance independent of such events.
MOTA (Multi Object Tracking Accuracy) \cite{bernardin2008evaluating} quantifies the proportion of correctly tracked instances by penalizing false negatives, false positives, and identity switches. Higher values indicate overall robustness, although the metric is known to emphasize detection quality more strongly than association consistency \cite{guan2025multi}.

While MOTA focuses on the correctness of object associations, MOTP (Multiple Object Tracking Precision) \cite{bernardin2008evaluating} evaluates the spatial accuracy of those associations. Specifically, MOTP measures the spatial alignment between predicted and ground-truth object locations across all matches. It reflects how precisely the tracker localizes objects, which is particularly relevant in cell tracking, where small spatial deviations can impact downstream analysis.

In addition, we used IDF1 \cite{ristani2016performance} which evaluates the temporal consistency of identity preservation by computing the harmonic mean of identity precision and recall. Unlike MOTA, IDF1 directly focuses on how well object trajectories are maintained over time, thereby providing complementary insights into association quality.

Lastly, the HOTA (Higher Order Tracking Accuracy) metric \cite{luiten2021hota} jointly evaluates detection and association quality as the geometric mean of detection accuracy (DetA) and association accuracy (AssA):
\begin{equation}
\begin{split}
    \text{HOTA}_\alpha &= \sqrt{\text{DetA}_\alpha * \text{AssA}_\alpha}
\end{split}
\label{eq:hota}
\end{equation}
Here, the detection accuracy (DetA) is the standard Jaccard index over predicted and ground truth detections, while the association accuracy (AssA) sums Jaccard indices over true positives, with each index computed only between the trajectories corresponding to that detection. This provides a unified measure of detection and association performance, while assessing the contributions of each component individually. Unlike CTC metrics, HOTA offers a balanced evaluation that avoids bias toward either detection or association errors alone.

While these MOT metrics provide a detailed view of raw tracking accuracy, it is important to note that, they do not explicitly account for cell-specific events such as mitosis. Consequently, they capture detection, association, and localization quality, but do not evaluate lineage tracking performance.

Regarding the hyperparameters for our evaluation, we used $\tau = 0.45$ for high-confidence detections and $\tau_{\text{low}} = 0.25$ for low-confidence detections. Additionally, we only considered objects that meet the distance and similarity thresholds of $\tau^{\text{dst}} = 50$ and $\tau^{\text{sim}} = 65$. To achieve a balanced weighing between spatial distances and similarity, we set $\lambda = 0.5$. Lastly, lost cells were retained for up to $n=5$ frames before being discarded from the memory bank. 

\subsection{Results}
\label{subsec:evaluation:results}
Direct comparison of the \name tracking accuracy with the one of existing methods is challenging as the existing approaches address different problem settings. Methods developed in the context of the CTC predominantly target single-channel recordings with constant signals, focusing on mitosis while neglecting apoptosis \cite{mavska2023cell}. Conversely, approaches that mention apoptosis often treat it implicitly, for example as cells vanishing from the field of view \cite{yazdi2024survey}, or rely on datasets without explicit apoptosis annotations \cite{akram2017cell}. In contrast, our framework jointly tracks cells, distinguishes between living and dead cells, and operates on multi-channel recordings with transient signals. This makes our framework applicable to more complex imaging scenarios and allows the extraction of richer biological insights. In the following, we present its performance using CTC and MOT metrics.

\subsubsection{CTC Evaluation}
\label{subsubsec:evaluation:results:ctc}
Regarding the CTC metrics (Table~\ref{table:evaluation:results}), \name achieved a DET score of $0.9395$, the LNK measure of $0.9156$ and a TRA score of $0.9364$. For reference, DET scores reported in the CTC challenge range from $0.876$ to $0.997$, LNK scores from $0.947$ to $1.000$, and TRA scores from $0.802$ to $0.994$ across different datasets and their best-performing methods. Although these values are not directly comparable due to substantial differences in the datasets and their difficulty, it is noteworthy that our method achieved scores within a similar range — except for the LNK score — even though it was evaluated on a considerably more challenging dataset.

\begin{figure}
    \centering
    \includegraphics[width=\columnwidth]{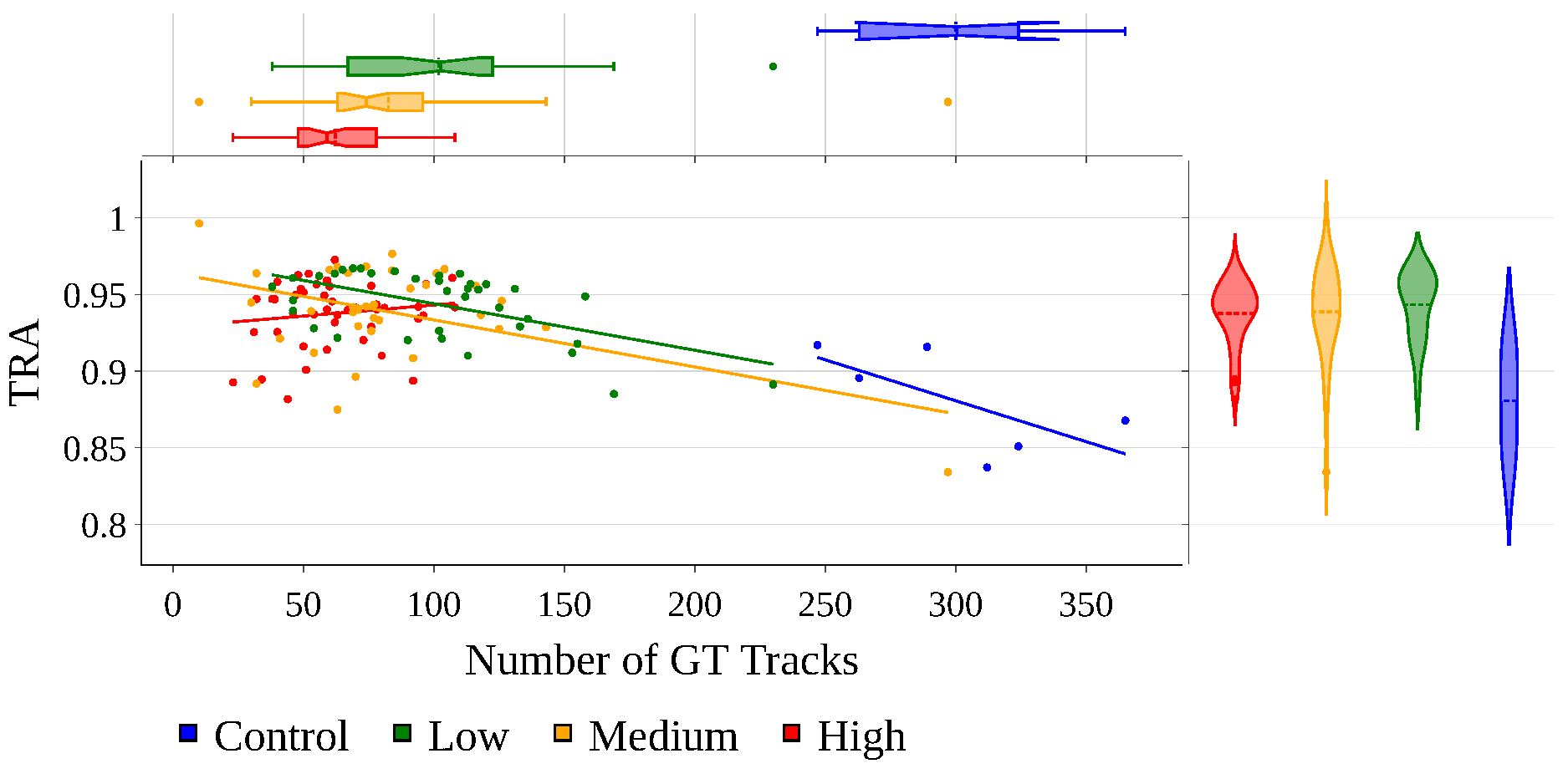}
    \caption{Performance of \name on 123 videos across control and three drug-dose groups, shown by TRA scores (↑ better). Top boxplots indicate the number of ground-truth tracks per group (video density), and right violin plots show TRA distributions.}
    \label{fig:performance_tracks_dose}
\end{figure}

Figure~\ref{fig:performance_tracks_dose} presents the performance of \name across different drug dose groups, evaluated using the TRA metric and their respective number of ground truth tracks. The results indicate that our approach achieved consistently high performance across the high-, medium-, and low-dose groups, which contain an average of $62$, $83$, and $102$ tracks per video, respectively. However, performance slightly declines as cell density increases, particularly in the control group. This group includes the fewest videos, yet exhibits the highest average track count per video ($\sim$300). Despite this, \name maintains strong tracking accuracy, with average TRA scores of $0.9434$, $0.9387$, and $0.9376$ for the low-, medium-, and high-dose groups, respectively.
\begin{table}[h!]
    \caption{Mean performance values of \name on the tracking videos, together with their standard deviations.}
    \centering
    \begin{tabular}{|c|cc|}
        \hline
         & \textbf{Metric} & \textbf{Score} \\\hline
        \multirow{3}{*}{\textbf{CTC Metrics}} & DET & $0.9395 \pm 0.0098$ \\
         & LNK & $0.9156 \pm 0.0179$ \\
         & TRA & $0.9364 \pm 0.0104$ \\
         \hline
        \multirow{6}{*}{\textbf{MOT Metrics}} & HOTA & $0.7670 \pm 0.0366$ \\
         & DetA & $0.7775 \pm 0.0328$ \\
         & AssA & $0.7579 \pm 0.0460$ \\
         & MOTA & $0.7963 \pm 0.0421$ \\
         & MOTP & $0.9253 \pm 0.0055$ \\
         & IDF1 & $0.8161 \pm 0.0412$ \\
        \hline
    \end{tabular}
    \label{table:evaluation:results}
\end{table}

\subsubsection{MOT Evaluation}
\label{subsubsec:evaluation:results:mot}
\name showed consistent MOT performance across all evaluated metrics (Table~\ref{table:evaluation:results}). For overall tracking accuracy, as measured by HOTA, it reached $0.767$, with detection accuracy (DetA) of $0.778$ and association accuracy (AssA) of $0.758$. The MOTA score of $0.796$ indicates that most cells were tracked correctly despite the challenging imaging conditions, although some errors from missed detections, false positives, or identity switches still occurred. In contrast, the MOTP score exceeded $0.925$, highlighting that once cells were detected, their positions were estimated with very high spatial precision. The IDF1 score of $0.816$ confirmed that object identities were maintained reliably in time, supporting the construction of accurate lineage trees and the detection of mitotic and apoptotic events. Taken together, these results suggest that \name excels at precise cell localization while occasionally facing difficulties in maintaining perfect temporal associations, an expected outcome in videos with transient fluorescent markers, heterogeneous signal quality, and frequent apoptotic events. 

\begin{table*}
    \centering
    \caption{Mean tracking performance (± standard deviation) of \name and ablation study, showing results with all features, without low-confidence detections, without Kalman Filter-based track interpolation, and without both.}
    \begin{tabular}{|c||c|c|c|}
    \hline
    Method & DET & LNK & TRA \\\hline
    \name & $\mathbf{0.9395 \pm 0.0098}$ & $\mathbf{0.9156 \pm 0.0179}$ & $\mathbf{0.9364 \pm 0.0104}$ \\\hline
    Only High-Conf. Detections & $0.9202 \pm 0.0153$ & $0.8794 \pm 0.0321$ & $0.9149 \pm 0.0170$ \\\hline
    No Kalman Filter & $0.9345 \pm 0.0132$ & $0.8981 \pm 0.0266$ & $0.9297 \pm 0.0144$ \\\hline
    None & $0.9083 \pm 0.0162$ & $0.8571 \pm 0.0337$ & $0.9017 \pm 0.0178$ \\\hline 
    \end{tabular}
    \label{tab:evaluation:ablations:features}
\end{table*}

\subsection{Ablations}
\label{subsec:evaluation:ablations}
In order to assess the contribution of various features or our method to the accuracy, we conducted ablation experiments by selectively omitting them (Table~\ref{tab:evaluation:ablations:features}). We considered four configurations: all features enabled, usage of only high-confidence detections, no track interpolation with Kalman Filter, and neither. In order to evaluate the different settings, we adhered to the CTC metrics only, since they are more suitable for capturing detection, tracking, and mitosis detection. Only using high-confidence detections led to a substantial drop in performance, with the mean scores for the metrics decreasing from $0.9395$ to $0.9202$ (DET), $0.9156$ to $0.8794$ (LNK), and from $0.9364$ to $0.9149$ (TRA). This underscores the importance of low-confidence detections as a fallback mechanism for handling hard-to-detect cells and improving matching accuracy. Additionally omitting track interpolation with the Kalman Filter further degraded performance, with DET, LNK, and TRA scores at $0.9083$, $0.8571$, and $0.9017$, respectively. By contrast, omitting interpolation alone resulted in only a modest performance decline (DET: $0.9345$, LNK: $0.8981$, TRA: $0.9297$), highlighting that low-confidence detections contribute most strongly to tracking performance.

\begin{figure}
    \centering
    \includegraphics[width=.9\columnwidth]{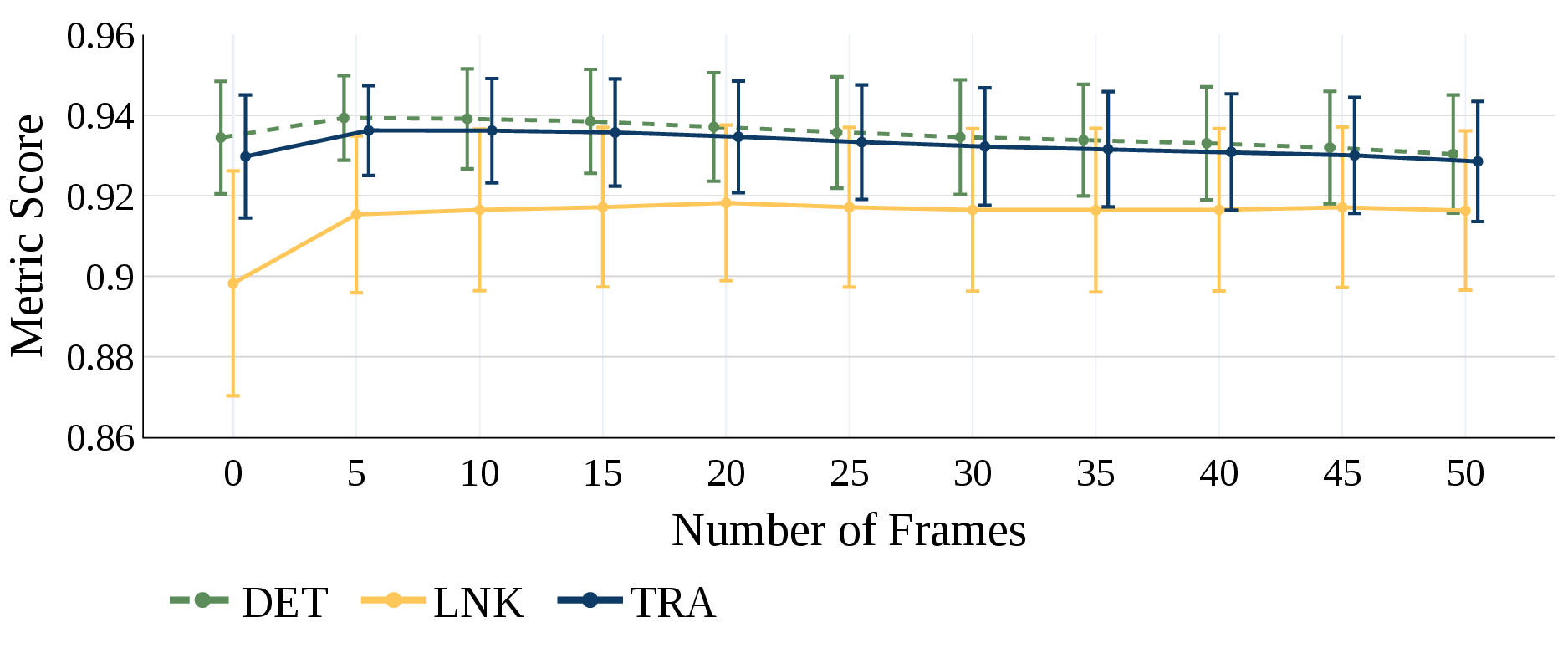}
    \caption{Performance on the CTC metrics for the tracking dataset when keeping lost cells in the memory bank for $n$ frames, including them in subsequent matchings.}
    \label{fig:ablations:patience}
\end{figure}

Next, we analyzed the effect of the maximum number of frames $n$ for which we retain lost cells in the memory bank (see Fig.~\ref{fig:ablations:patience}). Retaining cells for at least $5$ frames improved performance from a TRA score of $0.9297$ to $0.9364$, but extending the memory beyond $5$ frames yielded no further benefit. Notably, keeping cells beyond $10$ frames even led to a decline in TRA performance, likely because outdated candidates complicated the matching process. We observe the same effect of $n$ on DET. For the LNK metric, retaining cells for $5$ frames led to an initial improvement from $0.8981$ to $0.9156$, but extending the memory further had only a limited effect.

%% file: chapter/analysis.tex
\section{Analysis}
\label{sec:analysis}

We next applied \name to the entire videos, including unannotated frames, to enable a large-scale analysis of cell division behavior and inheritability. Specifically, we processed $154$ videos in total, $36$ each for the control-, low-, and medium-dose groups, and $46$ for the high-dose group. Across these datasets, we tracked $28{,}890$ individual cells, identifying $8{,}773$ cell divisions and $2{,}345$ cell deaths.

\subsection{Treatment Insights from Cell Tracking}
\label{subsec:analysis:insights}
\begin{figure*}[t!]
    \centering
    \includegraphics[width=0.93\textwidth]{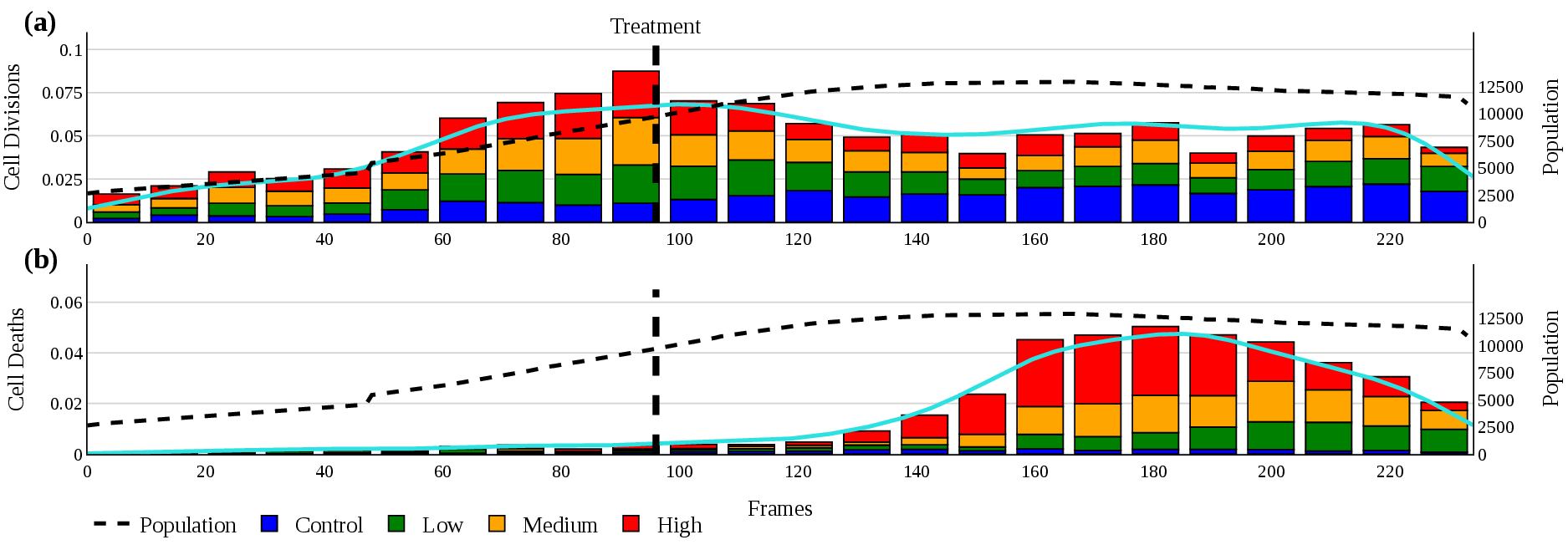}
    \caption{Number of cell division and death events per frame, normalized by cell counts and stratified by drug dosage groups. Frames are aggregated into bins of size 10. (a) shows the cell divisions and (b) the cell deaths. The overall cell population is shown on the secondary $y$-axis.}
    \label{fig:insights}
\end{figure*}

First, we employed our method to characterize cell population dynamics over time and to quantify the occurrence of cell divisions and cell deaths under different cisplatin dosages. Figure~\ref{fig:insights} summarizes the cell events by showing the frequency of cell divisions (Fig.~\ref{fig:insights}a) and cell deaths (Fig.~\ref{fig:insights}b) per frame. Prior to treatment (frame 96, corresponding to 48 hours), the mitotic rate increased steadily across all groups. At time of treatment, the division rate peaked and then declined sharply in all drug-exposed groups. The control group, by contrast, maintained a stable division rate that continued to increase after treatment. These observations indicate a clear dose-dependent suppressive effect of cisplatin on proliferation (Fig.~\ref{fig:insights}a).

Apoptosis rates remained low and stable before treatment. Approximately 40 frames ($\sim20$ hours) after cisplatin exposure, apoptosis began to increase across all treated groups, peaking between frames 180 and 190 ($\sim45$ hours post-treatment). This lag shows that the cytotoxic effect of cisplatin is delayed rather than immediate, likely due to the time that takes cells to attempt to correct the DNA-damage induced by the cisplatin \cite{gatti2002apoptosis}. Higher doses triggered an earlier and stronger apoptotic response in comparison to lower doses, reflecting a more pronounced cytotoxic effect. This dose-dependence is due to the fact that cells with high levels of damage do not try to repair their DNA and proceed directly to apoptosis \cite{price2006dependence,granada2020effects}.

The combined effect of reduced proliferation and increased apoptosis is visible in the total population trajectory (black dashed lines). Cell numbers grew until frame $150$ ($\sim75$ hours) but then gradually declined, consistent with an early dominance of proliferation followed by treatment-induced mortality.

In summary, these results illustrate the utility of our approach in quantifying cell dynamics and their response to treatment. By exposing the U-2 OS cell line to different doses of cisplatin, we unveiled the overall cell population behavior in response to a standard chemotherapeutic drug. Specifically, the method reveals, a dose-dependent suppression of mitosis, a delayed but robust induction of apoptosis, and the resulting turnover in net population growth.

\subsection{Cell Size Inheritability}
\label{subsec:analysis:inheritability}
For our further analysis, we restricted to cell trajectories that spanned at least 10 frames (5 hours) to exclude trajectories of cell briefly entering or exiting the field of view. For robustness, we constrained the analysis to the first five generations, as later generations contained too few divisions and cells for meaningful statistics.

Figure~\ref{fig:analysis:cell_size}(a) illustrates the inheritability of cell size across generations in the different dosage groups. We measure this inheritability as Pearson correlation of cell size between parent and descendant cells. The cell sizes were estimated from the bounding box measurements provided by \name. We excluded root cells (Generation 0) as the ancestor cells due to lack of information about their prior history. Correlations were weak from the start and declined steadily with each generation, reaching negligible levels by the fourth and fifth generation. The decline was faster in cisplatin-treated groups compared to controls. The weak and steadily declining ancestor–descendant correlations indicate that cell size is not strongly preserved across successive generations in our dataset. This suggests that lineage information alone is insufficient to explain variability in cell size over time. The faster loss of correlation observed in the cisplatin groups highlights that our method can capture treatment-dependent effects, reflecting increased heterogeneity in the cell populations under treatment compared to the controls.

\subsection{Sister Cell Synchronicity}
\label{subsec:analysis:synchronicity}
\begin{figure}[h!]
    \centering
    \includegraphics[width=.95\columnwidth]{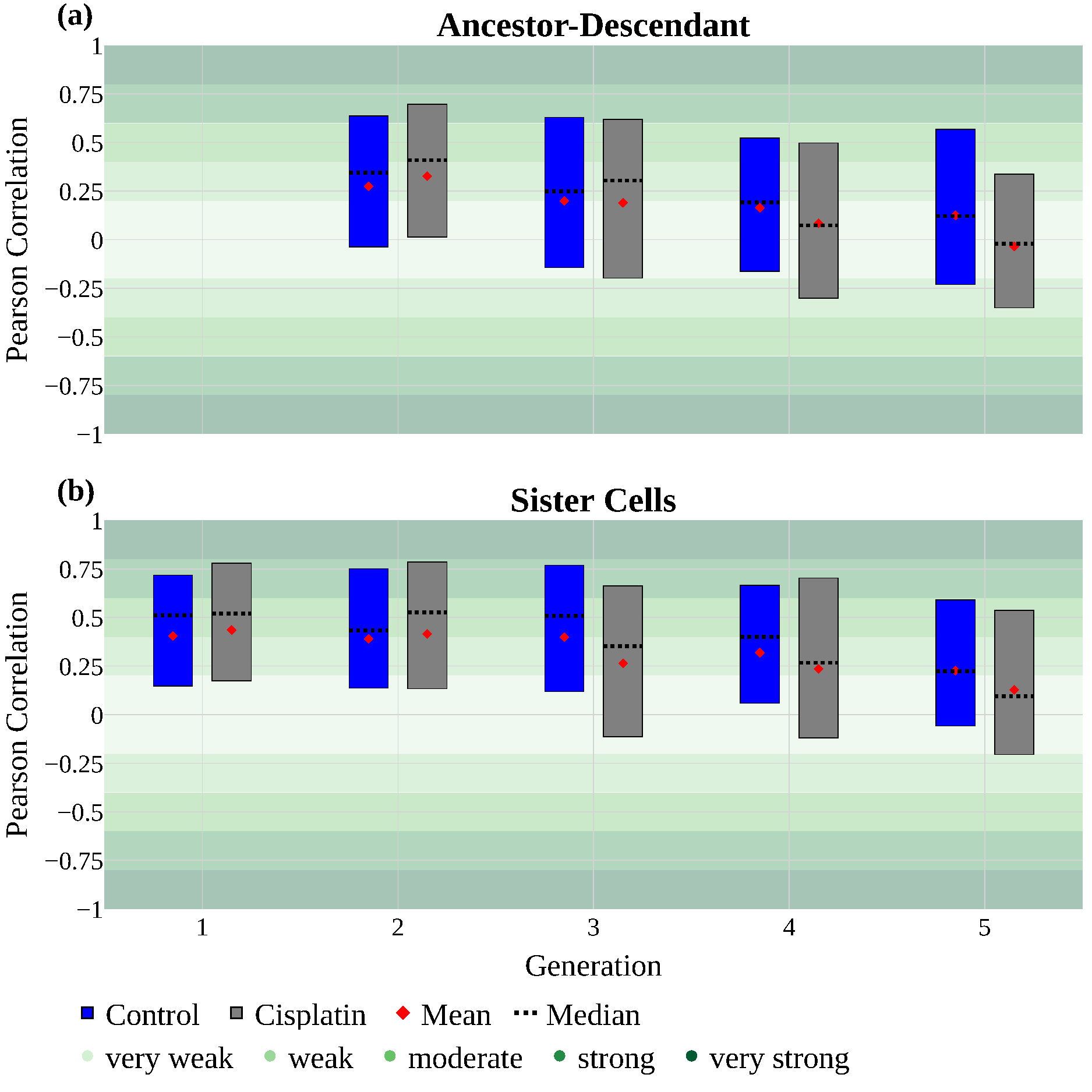}
    \caption{Pearson correlation of cell sizes across generations and within cisplatin-treated groups and the control group. (a) shows correlation of ancestor and descendant cell sizes and (b) of sister cell sizes.}
    \label{fig:analysis:cell_size}
\end{figure}

In addition to parent-descendant relationships, we examined the size correlation of sister cells (Fig. ~\ref{fig:analysis:cell_size}(b)). We computed the correlation of the cell sizes of sister cells, starting after division. Sister cell sizes remained positively correlated across all generations. Median correlations were moderate in the first generation and decreased gradually to weak levels by the fifth generation. Variability increased under cisplatin, with wider distributions and lower medians from generation three onward. The cells within the control groups showed more consistent correlations across generations, with a slower decline in correlation. The consistent positive correlations between sister cells indicate shared size characteristics immediately after division. The gradual decline across generations reflects increasing variability in the population over time. The wider distributions and faster decline of sister cell size correlation under cisplatin suggest that the treatment increases heterogeneity in cell populations compared to the more stable correlations observed in control populations.

\subsection{Cell Division}
\label{subsec:analysis:division}
\begin{figure}
    \centering
    \includegraphics[width=.95\columnwidth]{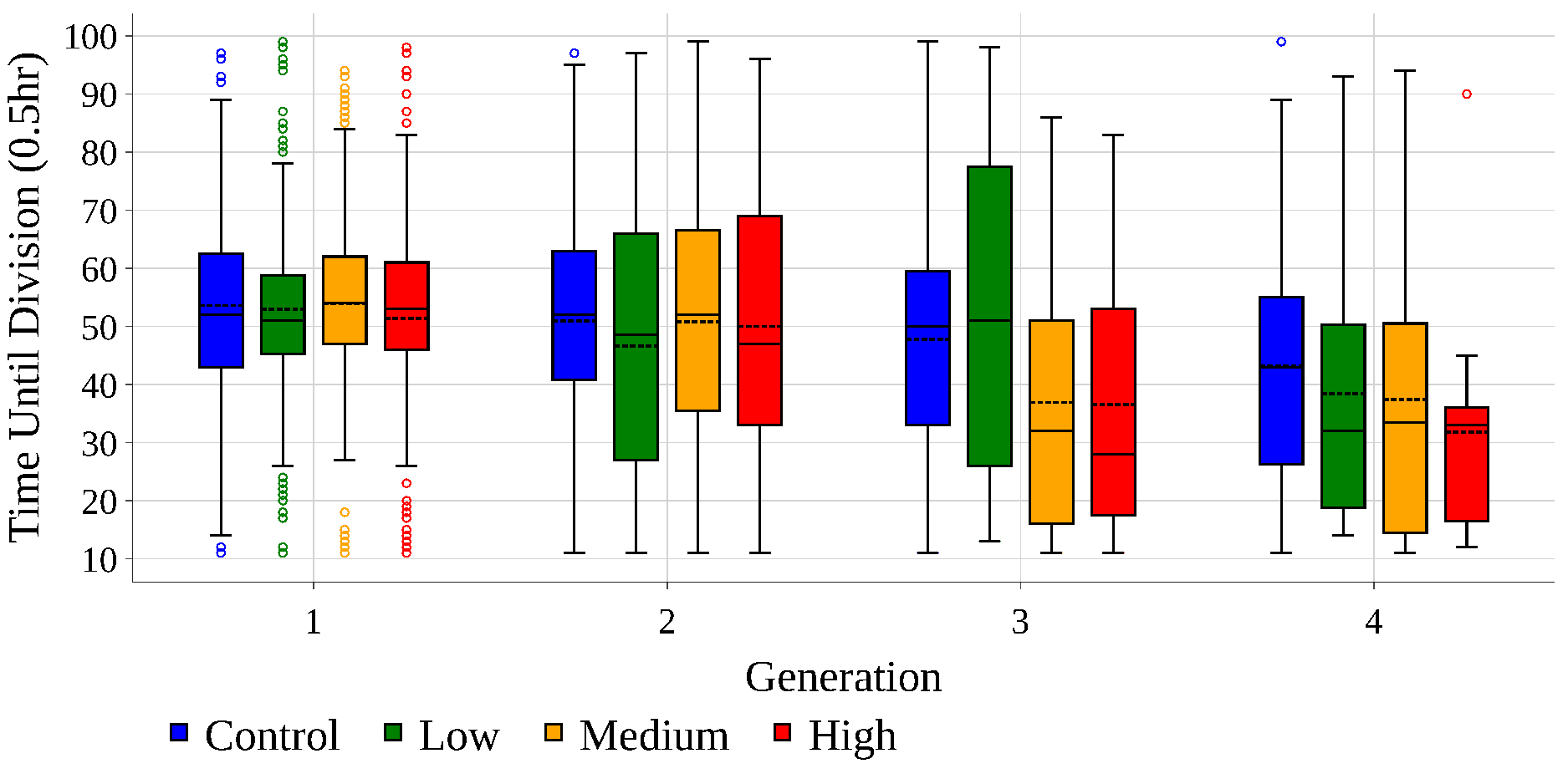}
    \caption{Time between cell divisions per dosage and per generation. Solid lines depict mean and dashed lines median.}
    \label{fig:analysis:time_until_division}
\end{figure}

We next analyzed the cell cycle length of dividing cells as the time between two consecutive divisions across dosage groups (Fig.~\ref{fig:analysis:time_until_division}). In this analysis, we excluded cells that took longer than $100$ frames ($50$ hours) to divide. Such divisions account for a small percentage of total divisions and likely represent cells that underwent a long process of repair from DNA damage after which they divided. Division times were comparable across dosage groups in the first generation, with mean values ranging from $25$ to $28$ hours. From the second generation onward, division times decreased monotonically in all groups, indicating progressively faster cell cycles in later generations. Overall, we observed substantial variability in interdivision times. High- and medium-dose groups exhibited earlier shifts toward shorter division times compared to controls.

\begin{figure}[t!]
    \centering
    \includegraphics[width=0.88\columnwidth]{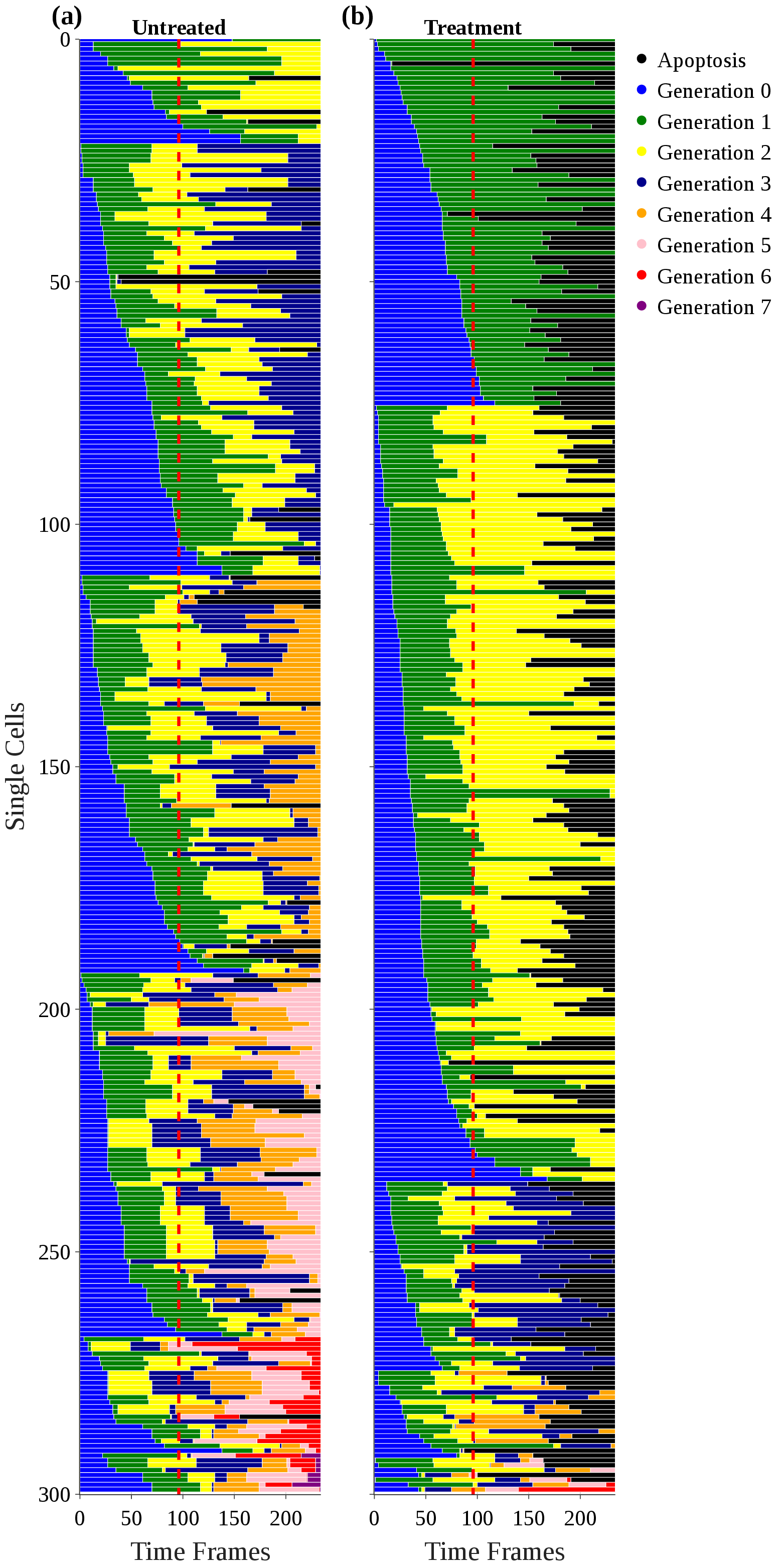}
    \caption{Division profiles of individual cells tracked across the entire dataset (234 frames, 117 hours). Panel (a) shows profiles for the control group, while panel (b) depicts the high-cisplatin group. Each row corresponds to a single cell, with mitotic and death events indicated by color transitions. Data were subsampled to 300 cells per group while preserving the distribution of division counts.}
    \label{fig:analysis:division_profile}
\end{figure}

Lastly, we analyzed the division profiles of different cell lineages (Fig.~\ref{fig:analysis:division_profile}). Lineage profiles were derived from trajectories that could be followed from birth until either the end of the experiment or apoptosis. For clearer visualization, we subsampled $300$ cells per panel while preserving the distribution of division counts. The control condition (Fig.~\ref{fig:analysis:division_profile}(a)) showed the longest and most complex lineage trees, with the highest number of successive divisions and the lowest fraction of apoptotic events. With increasing cisplatin concentration, division rates declined and mortality rates increased. Apoptosis was more frequent at higher doses, often occurring already in early generations, reflecting the strong cytotoxicity of cisplatin at elevated concentrations \cite{gatti2002apoptosis,price2006dependence}. This effect was particularly pronounced in the high-cisplatin group (Fig.~\ref{fig:analysis:division_profile}(b)).

While cells in the control and low-dose groups frequently reached later generations, medium and high doses strongly restricted lineage progression. Interestingly, the occurrence of cell division events in the presence of medium or high doses of cisplatin could indicate a degree of cisplatin resistance in a small subpopulation of cells, as described previously. Lastly, in lineages that underwent more than two divisions, the duration of the second generation (Generation 1) appeared relatively constant across conditions. This most likely reflects the fast clonal expansion that cells underwent in culture, highlighting the heritability of cell cycle times \cite{plambeck2022heritable}. 

%% file: chapter/conclusion.tex
\section{Discussion \& Conclusion}
\label{sec:conclusion}

In this paper, we introduced \name, a robust and efficient cell tracking approach for videos with multi-channel transient cell markers. By directly leveraging detection embeddings for cell matching, our method remains lightweight and easy to train, while supporting the detection of mitosis and apoptosis. To handle weak or noisy signals, \name integrates a two-stage matching strategy capable of handling low-quality detections. In addition, we implemented a short-term memory mechanism that stores temporarily lost or reappearing cells, ensuring continuous, gap-free trajectories. To evaluate performance, we used a challenging dataset comprising three transient fluorescent channels, in contrast to the single constant channel commonly used in prior work. The dataset presents additional challenges due to weak signals and cells that are difficult to detect. For a subset of videos, we annotated cell trajectories as well as mitotic and apoptotic events. Our labeled dataset includes $51{,}000$ trajectories, $18{,}000$ mitotic, and $3{,}500$ death events and represents a valuable resource for further development of tracking methods.

Our experimental evaluation highlighted both the strengths and current limitations of the proposed method. Regarding the CTC metrics, we observed comparable results for TRA and DET, consistent with the DetA and AssA components of the HOTA metric (Table~\ref{table:evaluation:results}), indicating a balanced contribution from detection and association performance. \name achieved strong results on videos containing small to medium numbers of cells but showed reduced performance on videos with numerous trajectories. This limitation may partly stem from the limited availability of densely populated videos in our training and evaluation sets. Future work should re-evaluate \name on datasets featuring denser cell populations and investigate whether the detector training on such data could mitigate these challenges.

In our ablation study (Table~\ref{tab:evaluation:ablations:features}), we highlighted the benefits and effectiveness of the combined method design and showed that the inclusion of low-confidence detections had the strongest impact on the tracking accuracy. Moreover, we demonstrated the utility of our memory bank (Fig.~\ref{fig:ablations:patience}), and showed that we only needed to store lost cells for up to $5$ frames to achieve the best performance. This suggests that most of the lost cells were recovered within fewer than $5$ frames, while those absent for longer have typically moved too far in space to be reliably matched.

Beyond evaluating tracking performance, we applied \name in a large-scale lineage analysis of cisplatin-treated cancer cells. In $154$ videos, we reconstructed trajectories of $29{,}000$ cells, detected nearly $8{,}800$ mitotic events and $2{,}300$ apoptotic events. This large-scale study highlighted well-known biological responses: inhibition of mitosis, dose-dependent induction of apoptosis, and the resulting turnover in net population growth. Furthermore, we showed that cell size is not reliably inherited across generations, while sister-cell size correlations persist but decline under treatment. Additionally, lineage progression is strongly curtailed at higher drug concentrations. 

Our analysis revealed cell lineages that showed robust cell cycle progression despite the presence of the chemotherapeutic drug. Such lineages may represent subpopulations with features of drug resistance, consistent with studies reporting mechanisms of cisplatin resistance in cancer cells \cite{amable2016cisplatin,zhang2025recent}. These lineages should be studied in depth as potential sources of tumor drug resistance. A study of their dynamics and temporal features in relationship to treatment timing could reveal important insights into the factors determining anticancer therapy efficacy. The complete lineage information, as well as the determination of outcomes of individual cells provided by \name, is key for enabling such studies, complementing and extending biological investigations into drug resistance. Our analysis underscores the utility of our method for analyzing dynamic cell populations and extracting treatment-dependent signatures from challenging imaging data.

Taken together, \name establishes a practical and comprehensive framework for studying cell populations under complex imaging conditions. By combining robust tracking with biologically meaningful endpoints, it enables quantitative studies of cell division, cell death, and lineage behavior. Future work should focus on improving cell detection, as higher-quality detections directly translate into more reliable associations and lineage reconstructions. Enhancing detection accuracy would not only simplify the matching process but also reduce error propagation across tracks. The application of our method to the study of cancer cell populations under treatment could help to identify and characterize in detail dynamic features involved in treatment resistance mechanisms.

%% file: literature.bib
@article{dai2022survey,
	title = {A survey of detection-based video multi-object tracking},
	volume = {75},
	journal = {Displays},
	author = {Dai, Yan and Hu, Ziyu and Zhang, Shuqi and Liu, Lianjun},
	year = {2022},
	pages = {102317},
}

@inproceedings{rezatofighi2019generalized,
	title = {Generalized intersection over union: a metric and a loss for bounding box regression},
	booktitle = {Proceedings of the {IEEE}/{CVF} conference on computer vision and pattern recognition ({CVPR})},
	author = {Rezatofighi, Hamid and Tsoi, Nathan and Gwak, JunYoung and Sadeghian, Amir and Reid, Ian and Savarese, Silvio},
	month = jun,
	year = {2019},
}

@inproceedings{lin2017focal,
	title = {Focal loss for dense object detection},
	booktitle = {Proceedings of the {IEEE} international conference on computer vision ({ICCV})},
	author = {Lin, Tsung-Yi and Goyal, Priya and Girshick, Ross and He, Kaiming and Dollar, Piotr},
	month = oct,
	year = {2017},
}

@article{yu2016cisplatin,
	title = {Cisplatin selects for stem-like cells in osteosarcoma by activating {Notch} signaling},
	volume = {7},
	number = {22},
	journal = {Oncotarget},
	author = {Yu, Ling and Fan, Zhengfu and Fang, Shuo and Yang, Jian and Gao, Tian and Simões, Bruno M and Eyre, Rachel and Guo, Weichun and Clarke, Robert B},
	year = {2016},
	pages = {33055--33068},
}

@article{advzemovic2025deep,
	title = {Deep {Learning}-{Based} {Multi}-{Object} {Tracking}: {A} {Comprehensive} {Survey} from {Foundations} to {State}-of-the-{Art}},
	journal = {arXiv:2506.13457},
	author = {Adžemović, Momir},
	year = {2025},
}

@article{stanojevic2024boosttrack,
	title = {{BoostTrack}: boosting the similarity measure and detection confidence for improved multiple object tracking},
	volume = {35},
	issn = {1432-1769},
	shorttitle = {{BoostTrack}},
	language = {en},
	number = {3},
	journal = {Machine Vision and Applications},
	author = {Stanojevic, Vukasin D. and Todorovic, Branimir T.},
	month = apr,
	year = {2024},
	pages = {53},
}

@inproceedings{bergmann2019tracking,
	title = {Tracking without bells and whistles},
	booktitle = {Proceedings of the {IEEE}/{CVF} international conference on computer vision ({ICCV})},
	author = {Bergmann, Philipp and Meinhardt, Tim and Leal-Taixe, Laura},
	month = oct,
	year = {2019},
}

@article{moen2019accurate,
	title = {Accurate cell tracking and lineage construction in live-cell imaging experiments with deep learning},
	journal = {bioRxiv : the preprint server for biology},
	author = {Moen, Erick and Borba, Enrico and Miller, Geneva and Schwartz, Morgan and Bannon, Dylan and Koe, Nora and Camplisson, Isabella and Kyme, Daniel and Pavelchek, Cole and Price, Tyler and {others}},
	year = {2019},
	note = {Publisher: Cold Spring Harbor Laboratory},
	pages = {803205},
}

@article{zhang2025recent,
	title = {Recent advances in the role of {circRNA} in cisplatin resistance in tumors},
	volume = {32},
	issn = {1476-5500},
	language = {en},
	number = {5},
	journal = {Cancer Gene Therapy},
	author = {Zhang, Jiawen and Yu, Qiwen and Zhu, Weijin and Sun, Xiaochun},
	month = may,
	year = {2025},
	pages = {497--506},
}

@article{amable2016cisplatin,
	title = {Cisplatin resistance and opportunities for precision medicine},
	volume = {106},
	issn = {1043-6618},
	journal = {Pharmacological Research},
	author = {Amable, Lauren},
	year = {2016},
	pages = {27--36},
}

@article{plambeck2022heritable,
	title = {Heritable changes in division speed accompany the diversification of single {T} cell fate},
	volume = {119},
	number = {9},
	journal = {Proceedings of the National Academy of Sciences},
	author = {Plambeck, Marten and Kazeroonian, Atefeh and Loeffler, Dirk and Kretschmer, Lorenz and Salinno, Ciro and Schroeder, Timm and Busch, Dirk H. and Flossdorf, Michael and Buchholz, Veit R.},
	year = {2022},
	pages = {e2116260119},
}

@article{price2006dependence,
	title = {Dependence of cisplatin-induced cell death in vitro and in vivo on cyclin-dependent kinase 2},
	volume = {17},
	issn = {1046-6673},
	language = {eng},
	number = {9},
	journal = {Journal of the American Society of Nephrology: JASN},
	author = {Price, Peter M. and Yu, Fang and Kaldis, Philipp and Aleem, Eiman and Nowak, Grazyna and Safirstein, Robert L. and Megyesi, Judit},
	month = sep,
	year = {2006},
	pmid = {16914540},
	pmcid = {PMC1698291},
	pages = {2434--2442},
}

@article{gatti2002apoptosis,
	title = {Apoptosis and growth arrest induced by platinum compounds in {U2}-{OS} cells reflect a specific {DNA} damage recognition associated with a different p53-mediated response},
	volume = {9},
	issn = {1476-5403},
	doi = {10.1038/sj.cdd.4401109},
	language = {en},
	number = {12},
	journal = {Cell Death \& Differentiation},
	author = {Gatti, L. and Supino, R. and Perego, P. and Pavesi, R. and Caserini, C. and Carenini, N. and Righetti, S. C. and Zuco, V. and Zunino, F.},
	month = dec,
	year = {2002},
	pages = {1352--1359},
}

@article{kuhn1955hungarian,
	title = {The {Hungarian} method for the assignment problem},
	volume = {2},
	number = {1-2},
	journal = {Naval research logistics quarterly},
	author = {Kuhn, Harold W},
	year = {1955},
	pages = {83--97},
}

@article{granada2020effects,
	title = {The effects of proliferation status and cell cycle phase on the responses of single cells to chemotherapy},
	volume = {31},
	number = {8},
	journal = {Molecular Biology of the Cell},
	author = {Granada, Adrián E and Jiménez, Alba and Stewart-Ornstein, Jacob and Blüthgen, Nils and Reber, Simone and Jambhekar, Ashwini and Lahav, Galit},
	year = {2020},
	pages = {845--857},
}

@article{mavska2023cell,
	title = {The cell tracking challenge: 10 years of objective benchmarking},
	volume = {20},
	number = {7},
	journal = {Nature Methods},
	author = {Maška, Martin and Ulman, Vladimı́r and Delgado-Rodriguez, Pablo and Gómez-de-Mariscal, Estibaliz and Nečasová, Tereza and Guerrero Peña, Fidel A and Ren, Tsang Ing and Meyerowitz, Elliot M and Scherr, Tim and Löffler, Katharina and {others}},
	year = {2023},
	pages = {1010--1020},
}

@inproceedings{wang2024smiletrack,
	title = {Smiletrack: {Similarity} learning for occlusion-aware multiple object tracking},
	volume = {38},
	booktitle = {Proceedings of the {AAAI} conference on artificial intelligence},
	author = {Wang, Yu-Hsiang and Hsieh, Jun-Wei and Chen, Ping-Yang and Chang, Ming-Ching and So, Hung-Hin and Li, Xin},
	year = {2024},
	pages = {5740--5748},
}

@article{li2022simpletrack,
	title = {Simpletrack: {Rethinking} and improving the jde approach for multi-object tracking},
	volume = {22},
	number = {15},
	journal = {Sensors},
	author = {Li, Jiaxin and Ding, Yan and Wei, Hua-Liang and Zhang, Yutong and Lin, Wenxiang},
	year = {2022},
	pages = {5863},
}

@article{turetken2016network,
	title = {Network flow integer programming to track elliptical cells in time-lapse sequences},
	volume = {36},
	number = {4},
	journal = {IEEE transactions on medical imaging},
	author = {Türetken, Engin and Wang, Xinchao and Becker, Carlos J and Haubold, Carsten and Fua, Pascal},
	year = {2016},
	pages = {942--951},
}

@article{guan2025multi,
	title = {Multi-object tracking review: retrospective and emerging trend},
	volume = {58},
	number = {8},
	journal = {Artificial Intelligence Review},
	author = {Guan, Zhiyu and Wang, Zhaofa and Zhang, Gan and Li, Luwei and Zhang, Miaomiao and Shi, Zhiping and Jiang, Na},
	year = {2025},
	pages = {235},
}

@article{luiten2021hota,
	title = {Hota: {A} higher order metric for evaluating multi-object tracking},
	volume = {129},
	number = {2},
	journal = {International journal of computer vision},
	author = {Luiten, Jonathon and Osep, Aljosa and Dendorfer, Patrick and Torr, Philip and Geiger, Andreas and Leal-Taixé, Laura and Leibe, Bastian},
	year = {2021},
	pages = {548--578},
}

@article{magnusson2014global,
	title = {Global linking of cell tracks using the {Viterbi} algorithm},
	volume = {34},
	number = {4},
	journal = {IEEE transactions on medical imaging},
	author = {Magnusson, Klas EG and Jaldén, Joakim and Gilbert, Penney M and Blau, Helen M},
	year = {2014},
	pages = {911--929},
}

@article{bernardin2008evaluating,
	title = {Evaluating multiple object tracking performance: the clear mot metrics},
	volume = {2008},
	number = {1},
	journal = {EURASIP Journal on Image and Video Processing},
	author = {Bernardin, Keni and Stiefelhagen, Rainer},
	year = {2008},
	pages = {246309},
}

@article{padfield2011coupled,
	title = {Coupled minimum-cost flow cell tracking for high-throughput quantitative analysis},
	volume = {15},
	number = {4},
	journal = {Medical image analysis},
	author = {Padfield, Dirk and Rittscher, Jens and Roysam, Badrinath},
	year = {2011},
	pages = {650--668},
}

@article{chatterjee2013cell,
	title = {Cell tracking in microscopic video using matching and linking of bipartite graphs},
	volume = {112},
	number = {3},
	journal = {Computer methods and programs in biomedicine},
	author = {Chatterjee, Rohit and Ghosh, Mayukh and Chowdhury, Ananda S and Ray, Nilanjan},
	year = {2013},
	pages = {422--431},
}

@article{matula2015cell,
	title = {Cell tracking accuracy measurement based on comparison of acyclic oriented graphs},
	volume = {10},
	number = {12},
	journal = {PloS one},
	author = {Matula, Pavel and Maška, Martin and Sorokin, Dmitry V and Matula, Petr and Ortiz-de-Solórzano, Carlos and Kozubek, Michal},
	year = {2015},
	pages = {e0144959},
}

@article{scherr2020cell,
	title = {Cell segmentation and tracking using {CNN}-based distance predictions and a graph-based matching strategy},
	volume = {15},
	number = {12},
	journal = {PLoS One},
	author = {Scherr, Tim and Löffler, Katharina and Böhland, Moritz and Mikut, Ralf},
	year = {2020},
	pages = {e0243219},
}

@article{yazdi2024survey,
	title = {A survey on automated cell tracking: challenges and solutions},
	volume = {83},
	number = {34},
	journal = {Multimedia Tools and Applications},
	author = {Yazdi, Reza and Khotanlou, Hassan},
	year = {2024},
	pages = {81511--81547},
}

@article{li2024lightweight,
	title = {A lightweight scheme of deep appearance extraction for robust online multi-object tracking},
	volume = {40},
	number = {3},
	journal = {The Visual Computer},
	author = {Li, Yi and Liu, Youyu and Zhou, Chuanen and Xu, Dezhang and Tao, Wanbao},
	year = {2024},
	pages = {2049--2065},
}

@article{ren2024rlm,
	title = {Rlm-tracking: online multi-pedestrian tracking supported by relative location mapping},
	volume = {15},
	number = {7},
	journal = {International Journal of Machine Learning and Cybernetics},
	author = {Ren, Kai and Hu, Chuanping and Xi, Hao},
	year = {2024},
	pages = {2881--2897},
}

@article{akram2017cell,
	title = {Cell tracking via proposal generation and selection},
	journal = {arXiv:1705.03386},
	author = {Akram, Saad Ullah and Kannala, Juho and Eklund, Lauri and Heikkilä, Janne},
	year = {2017},
}

@article{aharon2022bot,
	title = {{BoT}-{SORT}: {Robust} associations multi-pedestrian tracking},
	journal = {arXiv:2206.14651},
	author = {Aharon, Nir and Orfaig, Roy and Bobrovsky, Ben-Zion},
	year = {2022},
}

@inproceedings{ristani2016performance,
	title = {Performance measures and a data set for multi-target, multi-camera tracking},
	booktitle = {European conference on computer vision},
	publisher = {Springer},
	author = {Ristani, Ergys and Solera, Francesco and Zou, Roger and Cucchiara, Rita and Tomasi, Carlo},
	year = {2016},
	pages = {17--35},
}

@book{fukunaga2013introduction,
	title = {Introduction to statistical pattern recognition},
	publisher = {Elsevier},
	author = {Fukunaga, Keinosuke},
	year = {2013},
}

@inproceedings{yin2021center,
	title = {Center-based 3d object detection and tracking},
	booktitle = {Proceedings of the {IEEE}/{CVF} conference on computer vision and pattern recognition},
	author = {Yin, Tianwei and Zhou, Xingyi and Krahenbuhl, Philipp},
	year = {2021},
	pages = {11784--11793},
}

@inproceedings{huang2024deconfusetrack,
	title = {Deconfusetrack: {Dealing} with confusion for multi-object tracking},
	booktitle = {Proceedings of the {IEEE}/{CVF} conference on computer vision and pattern recognition},
	author = {Huang, Cheng and Han, Shoudong and He, Mengyu and Zheng, Wenbo and Wei, Yuhao},
	year = {2024},
	pages = {19290--19299},
}

@article{kalman1960new,
	title = {A new approach to linear filtering and prediction problems},
	author = {Kalman, Rudolf E},
	year = {1960},
}

@article{zhu2020deformable,
	title = {Deformable detr: {Deformable} transformers for end-to-end object detection},
	journal = {arXiv preprint arXiv:2010.04159},
	author = {Zhu, Xizhou and Su, Weijie and Lu, Lewei and Li, Bin and Wang, Xiaogang and Dai, Jifeng},
	year = {2020},
}

@inproceedings{zhang2022bytetrack,
	title = {Bytetrack: {Multi}-object tracking by associating every detection box},
	booktitle = {European conference on computer vision},
	publisher = {Springer},
	author = {Zhang, Yifu and Sun, Peize and Jiang, Yi and Yu, Dongdong and Weng, Fucheng and Yuan, Zehuan and Luo, Ping and Liu, Wenyu and Wang, Xinggang},
	year = {2022},
	pages = {1--21},
}

@inproceedings{yang2021multi,
	title = {Multi-object tracking with tracked object bounding box association},
	booktitle = {2021 {IEEE} international conference on multimedia \& expo workshops ({ICMEW})},
	publisher = {IEEE},
	author = {Yang, Nanyang and Wang, Yi and Chau, Lap-Pui},
	year = {2021},
	pages = {1--6},
}

@inproceedings{wojke2017simple,
	title = {Simple online and realtime tracking with a deep association metric},
	booktitle = {2017 {IEEE} international conference on image processing ({ICIP})},
	publisher = {IEEE},
	author = {Wojke, Nicolai and Bewley, Alex and Paulus, Dietrich},
	year = {2017},
	pages = {3645--3649},
}

@inproceedings{gallusser2024trackastra,
	title = {Trackastra: {Transformer}-based cell tracking for live-cell microscopy},
	booktitle = {European conference on computer vision},
	publisher = {Springer},
	author = {Gallusser, Benjamin and Weigert, Martin},
	year = {2024},
	pages = {467--484},
}

@inproceedings{bochinski2017high,
	title = {High-speed tracking-by-detection without using image information},
	booktitle = {2017 14th {IEEE} international conference on advanced video and signal based surveillance ({AVSS})},
	publisher = {IEEE},
	author = {Bochinski, Erik and Eiselein, Volker and Sikora, Thomas},
	year = {2017},
	pages = {1--6},
}

@inproceedings{bewley2016simple,
	title = {Simple online and realtime tracking},
	booktitle = {2016 {IEEE} international conference on image processing ({ICIP})},
	publisher = {Ieee},
	author = {Bewley, Alex and Ge, Zongyuan and Ott, Lionel and Ramos, Fabio and Upcroft, Ben},
	year = {2016},
	pages = {3464--3468},
}

@inproceedings{ben2022graph,
	title = {Graph neural network for cell tracking in microscopy videos},
	booktitle = {European conference on computer vision},
	publisher = {Springer},
	author = {Ben-Haim, Tal and Raviv, Tammy Riklin},
	year = {2022},
	pages = {610--626},
}
